\documentclass[sigconf]{acmart} 
\AtBeginDocument{%
  \providecommand\BibTeX{{%
    \normalfont B\kern-0.5em{\scshape i\kern-0.25em b}\kern-0.8em\TeX}}}

\setcopyright{none}
\copyrightyear{}
\acmDOI{}
\acmISBN{}
\acmConference[Data Science Seminar]{Data Science Seminar}{2020}{Uni Passau}

\usepackage{tikz}
\usetikzlibrary{arrows}
\graphicspath{{img/}}

\begin{document}
\title{After All, Only The Last Neuron Matters:  Comparing Multi-modal Fusion Functions for Scene Graph Generation}

\author{Mohamed Karim BELAID}
\email{belad01@ads.uni-passau.de}
\affiliation{%
  \institution{University of Passau}
  \city{Passau}
  \state{Germany}
  \postcode{94032}
}

\begin{abstract}
From object segmentation to word vector representations, Scene Graph Generation (SGG) became a complex task built upon numerous research results. In this paper, we focus on the last module of this model: the fusion function. The role of this latter is to combine three hidden states. 
We perform an ablation test in order to compare different implementations. First, we reproduce the state-of-the-art results using SUM, and GATE functions. Then we expand the original solution by adding more model-agnostic functions: an adapted version of DIST and a mixture between MFB and GATE. 
On the basis of the state-of-the-art configuration, DIST performed the best \textit{Recall @ K} which makes it now part of the state-of-the-art.
\end{abstract}

\maketitle

\section{Introduction and motivation}
A Scene Graph Generation (SGG) model is a computer vision method that allows computers to describe the object interactions inside an image. The input of the model is an image representing a scene. The output is a graph describing the scene. The graph nodes represent the objects detected in the image and the edges represent the relationship between these objects.

\begin{figure}[tb]
  \centering
  \includegraphics[width=1\linewidth]{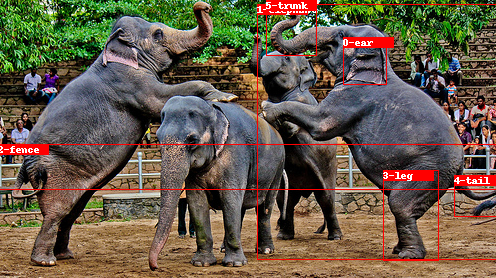}
  \caption{Example of input image, with ground truth object label and bounding boxes, from the VG dataset \cite{krishna2017visual}.}
  \label{fig:intro_ex}
\end{figure}

\begin{figure}[tb]
  \centering
  \includegraphics[width=1\linewidth]{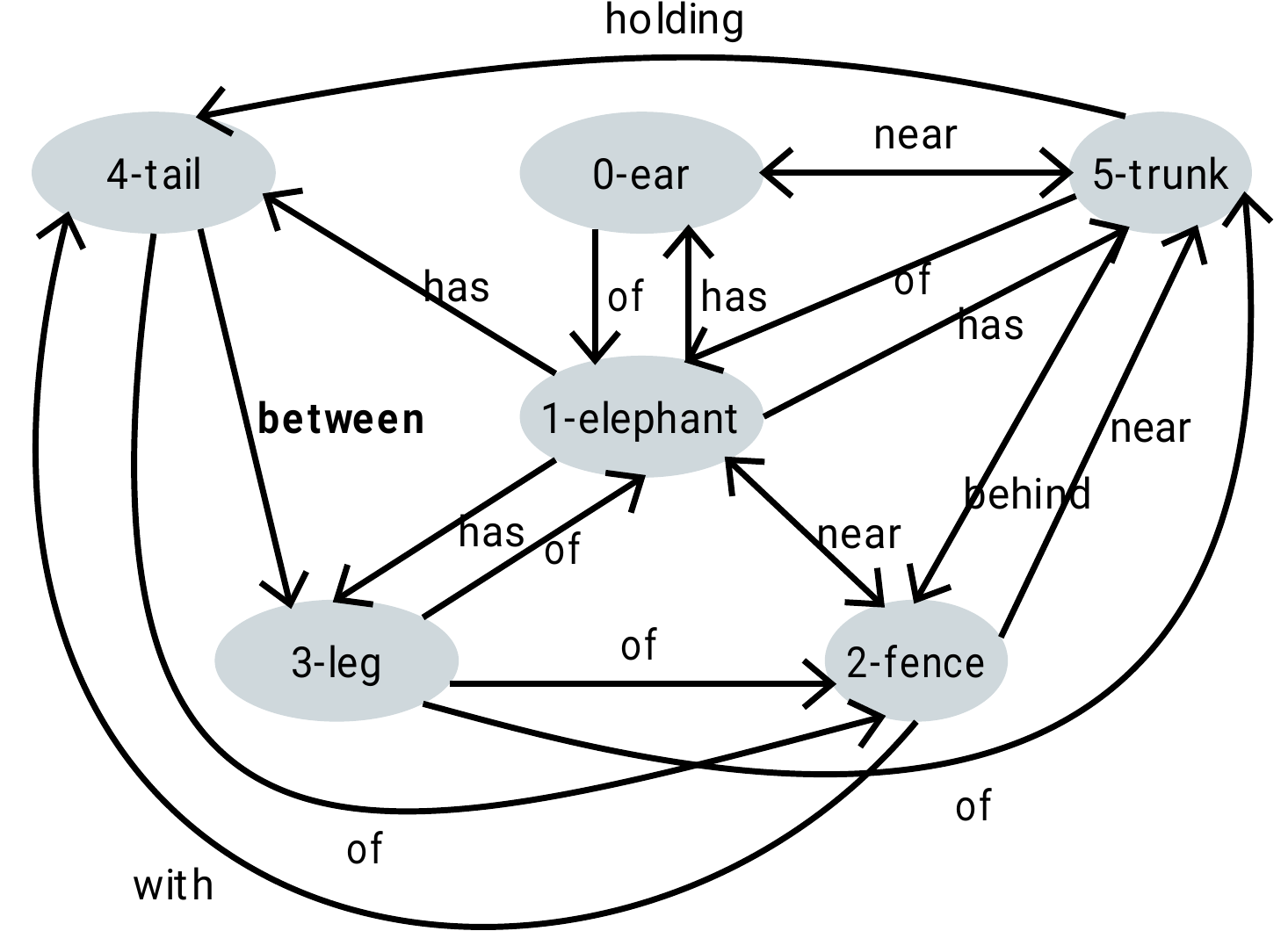}
  \caption{Example of a prediction of the predicates based on the ground truth object label and bounding boxes.}
  \label{fig:intro_ex_pred}
\end{figure}

Let us consider Figure~\ref{fig:intro_ex} as a basic example of an input and Figure~\ref{fig:intro_ex_pred} as an example of model prediction. The workflow of an SGG model is as follows: first, the objects in the image are detected (using for example a pre-trained neural net.). Also, their respective bounding box is defined. Second, the model extracts the predicates linking each pair of objects. To this end, the model relies on the object labels, two hidden states representing each object, and one hidden state representing the union between both objects. Third, the bias in the language is eliminated by applying Total Direct Effect (TDE), explained later on. 

What motivates us to work on the fusion function is that the most recent state-of-the-art SGG model was published recently \cite{tang2020unbiased}. And improving the fusion function is still to be done.


In this research, we study the fusion function used to merge all obtained hidden states. First, we reproduce the state-of-the-art result made with SUM and GATE functions. Then, we introduce new functions inspired by the Visual Question Answering domain. This allows us to identify the most important rules to obtain the best fusion function and thus the best SGG model.\footnote{We make our evaluation tools and resources, as well as the created state-of-the-art fusion functions available under open licenses from: \url{https://github.com/Karim-53/SGG}}
We first describe SUM and GATE methods in Section~\ref{sec:background}. Then, we present the new functions in Section~\ref{sec:method} and results in Section~\ref{sec:results} before concluding in Section~\ref{sec:conclusions}.

\section{Baseline SGG}
The Fusion Function receives 3 hidden states as input. To better understand the importance of the Fusion Function, let us embark on a long journey through the baseline neural network:\newline
\textbf{Image segmentation} Equation~\ref{eq:baseline1}. The first module identifies the candidate object labels ($l_i$)  in the image and their respective bounding box ($b_i$). We rely on ``Faster RCNN'' algorithm \cite{ren2015faster} which is itself quite complex (ConvNet + Region proposal network + MLP).\newline
\textbf{Visual contexts} Equation~\ref{eq:baseline2}. For each object $i$, the initial image is cropped around its bounding box $b_i$. So that the extracted hidden state $x_i$ represents the targeted object.\newline
\textbf{Object detection} Equation~\ref{eq:baseline3}. This module gets its input from the previous node. The goal is to identify the final object labels $\{z_i\}_n$.\newline 
\textbf{Object feature} Equation~\ref{eq:baseline4}. Starting from this module, the hidden states are processed in pairs in order to infer the predicate linking them. For example, this module gets as input the pair ($x_i$, $x_j$), the visual context of two objects. It outputs one hidden state describing the join representation. It is the first input of the fusion function.\newline 
\textbf{Object Class} Equation~\ref{eq:baseline5}. This module predicts the predicate label using only the subject and object labels. If Object~1 is a person and Object~2 is a chair, what would be the predicate linking these two? One would say "Person~1 is \underline{sitting~on} Chair 1" is the most probable solution, independently of the image. This inference does not depend on a computer vision algorithm, but only on the bias in the language. This bias will be eliminated later on using TDE.\newline
\textbf{Visual Context} Equation~\ref{eq:baseline6}. This hidden state encodes the visual context of one bounding box, the one that surrounds both targeted objects. In contrast, $\{x_{ij}\}$ does not encode the area between the two objects.\newline
\textbf{Biased predicate prediction} Equation~\ref{eq:baseline7}. The relation between the two objects is predicted by merging the three obtained hidden states. The Fusion function is used for this prediction. Their implementation will be detailed later on. The output $\{p_{ij}\}$ is the probability distribution of the predicates.\newline
\textbf{Final predicate prediction} Equation~\ref{eq:baseline8} and \ref{eq:baseline9} apply the Total Direct Effect (TDE). These last operations eliminate the bias due to the language. The bias is eliminated by calculating the difference between $p_{ij}$ the probability distribution of the predicate and $p_{ij | \bar{x}}$ the probability distribution of the predicates given no visual contexts $x_i$ and $x_j$. Indeed, without the visual contexts, $z_i$ and $z_j$ will represent only the bias. Thus, the prediction $p_{ij | \bar{x}}$ will be based only on the bias. 
In practice, $x$ is replaced by $\bar{x}$, the average hidden state. Also, one could replace it simply by zeros. 
\begin{equation}
    \text{Image} \mapsto n, \{l_i\}_n, \{b_i\}_n, \{r_i\}_n, M \label{eq:baseline1}\\
\end{equation}
$n$: the number of objects in the image.\newline
$\{l_i\}_n$: tentative object labels.\newline
$\{b_i\}_n$: bounding box. \newline
$\{r_i\}_n$: RoIAlign features \cite{he2017mask}.\newline
$M$: feature map of the entire image.
\begin{align}
(r_i , b_i , l_i) & \mapsto {x_i} & \forall i \leq n
\label{eq:baseline2}\\
x_i & \mapsto z_i & \forall i \leq n\label{eq:baseline3}
\end{align}
$x_i$: hidden state that represents Object $i$ (encoded visual contexts).\newline
$z_i$: fine-tuned label of Object $i$ (distribution).
\begin{align}
( x_i, x_j ) & \mapsto x_{ij} & \forall i,j \leq n\label{eq:baseline4}\\
( z_i, z_j ) & \mapsto  z_{ij} & \forall  i,j \leq n\label{eq:baseline5}\\
(b_i \cup b_j, M) & \mapsto v_{ij} & \forall  i,j \leq n\label{eq:baseline6}
\end{align}
$x_{ij}$: join features representation of Object $i$ and $j$ (Object Feature Input for SGG).\newline
$z_{ij}$: predicate class (distribution) linking Obj. $i$ and $j$ inferred using only the object labels (Object Class Input for SGG).\newline
$b_i \cup b_j$: Union between the two bounding boxes.\newline
$v_{ij}$: (Visual Context Input for SGG).
\begin{align}
(x_{ij}, z_{ij}, v_{ij}) & \mapsto p_{ij} & \forall  i,j \leq n\label{eq:baseline7}\\
(\bar{x}, z_{ij}, v_{ij}) & \mapsto p_{ij | \bar{x}} & \forall  i,j \leq n\label{eq:baseline8}\\
p_{ij\ \text{TDE}} & = p_{ij} - p_{ij | \bar{x}} & \forall  i,j \leq n\label{eq:baseline9}
\end{align}
$p_{ij}$: biased Predicate.\newline
$\bar{x}$: Average features representation.\newline
$p_{ij | \bar{x}}$: biased Predicate (distribution).\newline
$p_{ij\ \text{TDE}}$: unbiased predicate. Final output of the model.

In this section, we have explained the algorithms in interaction with the fusion function: 3 neural networks provide the 3 inputs of the function. Afterward, TDE is calculated in one mathematical operation. The final output is the set of subject-object-predicate tuples ( $z_i$, $p_{ij\ \text{TDE}}$, $z_j$). Finally, the reader should be aware that the above framework is not applicable to all SGG. For example, some are based on Attention Models  \cite{belaid2019comparison} 
like \cite{yu2017multi, lu2016hierarchical}.

\section{Background}\label{sec:background} 
From object detection 
to image captioning
, machine learning and computer vision proved their exclusive potential in solving complex tasks. In this research, we tackle a more human level task: scene graph generation, a well popular task nowadays \cite{newell2017pixels,zellers2018neural,chen2019knowledge,tang2020unbiased}.
SGG is a multi-modal framework. The focus of this paper is the last fusion function, because of its proved high impact on the performance on SGG-related tasks \cite{yu2017multi}.

The baseline SGG model --- used in this research --- was implemented by Tang et al. \cite{tang2020unbiased} in March 2020 as it represents the most recent state-of-the-art configuration. The latter research already tests two fusion functions SUM and GATE, as stated in Equations \ref{eq:sum} and \ref{eq:gate}.
\begin{align}
SUM: p_{ij} &= W_x x_{ij} + W_v v_{ij} + z_{ij} \label{eq:sum}\\
GATE: p_{ij} &= W_r x_{ij} \cdot \sigma \left ( W_x x_{ij} + W_v v_{ij} + z_{ij} \right ) \label{eq:gate}
\end{align}
SUM is based on sum operation, and linear projections (multiplication by a matrix of coefficients). GATE utilizes the same term as SUM, and process it using a Sigmoid function. GATE also applies a dot product which is an element wise multiplication between both terms. This operator is also known as the Hadamard operator in other papers \cite{qi2020two}). Mainly, SUM and GATE are very popular among SGG models \cite{zhou2015simple, lu2016hierarchical, tang2020unbiased, qi2020two}.

One way to improve the fusion function is to add one term that measure the difference between the projections. For example, we can use the squared euclidean distance between two hidden states~\cite{zhang2018learning}. We refer to this fusion function as DIST and it is formalized mathematically in Equation~\ref{eq:dist}. A more advanced approach implements a Multi-modal Factorized Bilinear pooling (MFB)~\cite{yu2017multi}. Comparing to the former methods, MFB is the unique fusion function that expand the number of dimensions before cutting it again using sum pooling. This approach involves more weights and a thorough learning of the co-attention patterns that link each pair of inputs. Equation~\ref{eq:mfb} summarizes the implementation of MFB. A cascading of MFB blocs is named Multi-modal Factorized High-order Pooling (MFH) \cite{yu2018beyond}. This extension of MFB showed better results. Finally, the latter three approaches target Visual Question Answering (VQA). Thus, all functions have only two inputs: the question and the image.

\begin{align}
\text{DIST:}\ x \diamond  y &= \text{ReLU} \left ( W_x\ x \cdot W_y\ y \right ) - \left ( W_x\ x - W_y\ y \right ) ^2\label{eq:dist}\\
\text{MFB:}\ x \diamond  y &= \text{SumPool} \left ( W_x\ x \cdot W_y\ y \right )\label{eq:mfb}
\end{align}


\section{Methods}\label{sec:method}
In addition to the existing fusion function --- SUM and GATE --- we propose a modified version of the DIST function inspired by \cite{zhang2018learning}. Equation~\ref{eq:dist2} represents our implementation. Note that the main difference concerns the number of inputs: our implementation emphasis the difference between $x_{ij}$ (the join features representation of the subject $i$ and object $j$) and $v_{ij}$ (the visual context). Theoretically, it will less efficient to subtract $z_{ij}$ from $x_{ij}$ or $v_{ij}$, since $z_{ij}$ is inferred from the object classes while the two other states are inferred from the image.

In addition, we propose another fusion function. It is a combination between MFB and GATE as shown in Equation~\ref{eq:mfb7}. The first term represents the co-attention module of the MFB. It uses also $x_{ij}$ and $v_{ij}$, as for DIST. The second term represents the core of the fusion function since it merges the three hidden states. Finally, the element-wise dot product represents the GATE function as seen in the background.

\begin{align}
p_{ij} &= \sigma ( W_x x_{ij} + W_v v_{ij} + z_{ij} ) - \sigma( W_x x_{ij} - W_v v_{ij} )^2
\label{eq:dist2}\\
p_{ij} &= \sigma ( W_x x_{ij} \cdot W_v v_{ij} ) \cdot \sigma( W_x x_{ij} + W_v v_{ij} + z_{ij} )\label{eq:mfb7}
\end{align}

Similar candidate functions are tested, but not reported in this paper because they showed less comparative results.

\section{Experiments}\label{sec:experiments} 
\subsection{Experimental setup}
The used SGG model is built upon many other modules with their own set of parameters. Table \ref{tab:Results_setup} condenses the most important ones in order to reproduce the results.\footnote{For the complete list of parameters, please consult the following file\\ ``configs/e2e\_relation\_X\_101\_32\_8\_FPN\_1x.yaml'' in the GitHub repository.} Specifying the task to be ``Predicate Classification'' means that the model will obtain the ground truth object labels and bounding boxes as additional input. 
Note that the batch size is higher than in the original paper, which might explain why the results improved slightly. Concerning the maximum number of iterations, note that the training is monitored with an Early Stopping method. Thus, the training usually stops before reaching this threshold.

\begin{table}[bt]
  \caption{Experimental setup.}
  \label{tab:Results_setup} 
  \begin{tabular}{lc}
    \toprule
    Parameters&Value\\
    \midrule
    Task				& Predicate Classification\\
Predictor			& Motifs \cite{zellers2018neural} \\
Batch size 			& 64\\
maximum iteration 		& $50 000$ \\
Validation Period 		& $2 000$ \\
Pretrained word embedding 	& Glove \cite{pennington2014glove} \\ 
Pretrained image segmentation	& Faster R-CNN \cite{ren2015faster} \\
  \bottomrule
\end{tabular}
\end{table}

The implementation is made with Pytorch and it is compatible with parallel computing. Thus, the batch is distributed over 8 GPUs of type \textit{RTX 2080 ti}. The batch size is 8 per GPU. The distributed computing allowed us to train the model in 7 to 10 hours.

\subsection{Dataset}\label{dataset}

The used dataset was published in 2017 under the name of Visual Genome (VG) \cite{krishna2017visual}. The dataset includes over 100K images. With 18 predicates per image on average, VG is the densest and largest dataset for SGG. Of course, one could think about MS-COCO \cite{lin2014microsoft} with its 300k captioned images. But, the format of the five captions per image is not suitable for the Predicate Classification task. Nevertheless, the format of this dataset fits the Sentence-to-Graph Retrieval task.
For the image preprocessing, we follow the same instructions as the baseline implementation \cite{tang2020unbiased}: the longest side of each image is scaled to 1000 pixels.

The main issue we have to look at, concerning VG dataset, is the unbalance in the object and predicate categories. Indeed, \{\textit{on, has, wearing, of, in}\} are the five most frequent predicates. They account for nearly 75\% of the used predicates. For this reason, many research like \cite{xu2017scene, zellers2018neural, tang2019learning, chen2019counterfactual, tang2020unbiased} limited the ground truth graphs to the 150 most frequent objects and the 50 most frequent predicates.

\subsection{Metrics} 
SGG is a complex task. Each proposed solution has its own pros and cons. Hopefully, we can measure the different advantages using the following metrics.

\begin{figure*} 
  \centering
  \includegraphics[width=1\linewidth]{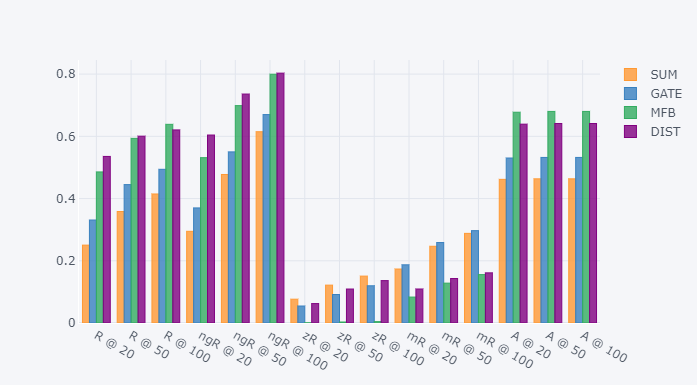}
  \caption{Comparing the results of different fusion functions on the predicate classification task.}
  \label{fig:results}
\end{figure*}

\subsubsection{mAP}
mAP evaluates only the object detection and precisely the quality of the bounding box. In this research work, we limit the experiment to predicate detection i.e. the object labels and the bounding boxes are part of the input. Thus, this metric is not relevant.

\subsubsection{Recall@K (R@K)}
R@K is a metric inspired by retrieval tasks. For each pair of subject-object in the image, the model proposes one candidate predicates, the most probable one. "Recall~At~K" metric focuses only on the K tuples (subject, predicate, object) with the highest confidence. Precisely, it is the ratio between the number of correct predictions and K. 
R@K is one of the most popular metrics in SGG. First proposed by Lu et al. \cite{lu2016visual} in 2016, this metric fit perfectly any dataset that is partially labeled. Indeed, the ground truth annotation of VG does not include all relationship predicates and the model can still infer a correct relationship that is not included in the ground truth. 

\subsubsection{No Graph Constraint Recall@K (ngR@K)}
ngR@K \cite{newell2017pixels, zellers2018neural} is suggested in 2017 and represents an expansion of the R@K metric. For each pair of subject-object in the image, the model proposes a list of candidate predicates with a different confidence level.
The output of the SGG model is therefore a multigraph.
ngR@K is more adapted to the SGG task since certain scenes can be described with more than one relation. For example, one could say: the Car is \underline{parked on} the Street, but also the Car is \underline{on} the Street.

\subsubsection{Mean Recall@K (mR@K)}
mR@K \cite{tang2019learning, chen2019knowledge} is proposed in 2019 and deals with the problem of bias in VG dataset (see Section~\ref{dataset}). 
mR@K is the average R@K over all predicates. Therefore, this metric is independent of the frequency of each predicate in the available ground truth.

\subsubsection{Accuracy@K Accuracy (A@K)}
A@K \cite{zhang2019graphical} metric is used when the K ground-truth pairs (subject, object) are provided to the model as input. Thus, this metric can be used only for the Predicate Classification task and Scene Graph Classification task.

Many other metrics can be explored --- like the Recall@K for each predicate, the Zero-Shot Recall@K (zR@K) \cite{lu2016visual}, and the  Sentence-to-Graph Retrieval (S2G) \cite{tang2020unbiased} --- but which are less relevant for this paper. 
In conclusion, each metric evaluates a specific step of the workflow of SGG which enables us to identify the pro and cons of each model.

\section{Results and discussion}\label{sec:results}
Figure~\ref{fig:results} shows the performance of each fusion function using all 5 metrics. 
First, we notice that the reproduced results (SUM and GATE) are slightly different from the one reported in the source paper \cite{tang2020unbiased}. It is due to two main reasons: the increase of the batch size, as explained above, and the optimization of the implementation, reported by the author after publishing the source paper.

Depending on the real word application, one could choose the metric that describes better the final goal. DIST performed best on \textit{R@K} and \textit{ngR@K}. SUM, the simplest function, performed best on \textit{zR@K}. The proposed functions were not able to beat the state-of-the-art implementation on the \textit{mR@K}: GATE still has the best score. Finally, A mixture between MFB and GATE obtained the best \textit{A@K}.


This study has potential limitations. First, we did not provide any statistics on the results (the mean and standard deviation of the scores). This is due to the long training time: we need about 7 to 10 hours of training per run and per fusion function. Second, we used some pretrained components, like the Faster R-CNN, in order to accelerate the training. It would be more veracious to train the model end-to-end during each run to get more accurate measures on the influence of the fusion function. Third, we provided the object classes and bounding boxes as input to the model. Thus, this work focused on a specific task: predicate classification.

\section{Conclusions}\label{sec:conclusions}
We studied the problem of SGG using TDE. TDE operation requires the calculation of specific hidden states. Hence, the necessity to implement a fusion function, responsible for merging three flows of hidden states. We reproduced the results of the SUM and GATE functions. Then, we adapted the DIST fusion function from the VQA domain 
. The obtained model achieved the best \textit{Recall @ K} and \textit{ng-Recall @ K} score. Finally, we created a mixture between the MFB and GATE function. This latter model obtained the best \textit{Accuracy @ K}. To conclude, each tested fusion function performs better depending on the final application of the SGG model.

In the future, we would like to prove that DIST also achieves state-of-the-art results on similar tasks like Scene Graph Classification and Scene Graph Detection.

\clearpage
\begin{acks}
I would like to thank the Chair of Data Science (Passau University) for providing continuous support which offers the best working flow, especially for the hardware support.
This paper and the research behind it would not have been possible without the exceptional support of my supervisors, Mr.~Julian~Stier, Mr.~J\"org~Schl\"otterer and Prof.~Dr.~Michael \mbox{Granitzer}. I~would like to show my gratitude to Dorra Elmekki and the anonymous reviewers, for their comments on an earlier version of the paper. Although any errors are our own and should not tarnish the reputations of these esteemed persons.
\end{acks}


\bibliographystyle{ACM-Reference-Format}
\bibliography{bib}
\end{document}